\documentclass[runningheads]{llncs}
\usepackage[T1]{fontenc}
\usepackage{hyperref}
\usepackage{cite}

\usepackage{amsmath,amssymb,amsfonts,amsthm}
\usepackage{algorithmic}
\usepackage{graphicx}
\usepackage{textcomp}
\usepackage{xcolor, soul}
\usepackage{url}
\usepackage{array,booktabs}
\usepackage{framed}
\usepackage{hyperref}
\usepackage{float}
\usepackage{longtable}

\usepackage[capitalize]{cleveref}
\newcommand {\otoprule}{\midrule [\heavyrulewidth]} 
\newcolumntype {+}{ >{\global\let\currentrowstyle\relax}}
\newcolumntype {^}{ >{\currentrowstyle }}
 \newcommand {\rowstyle}[1]{\gdef\currentrowstyle{#1} %
 #1\ignorespaces
 }
\newcolumntype{L}[1]{>{\raggedright\arraybackslash}p{#1}}
\newcommand{\tabhead}{\rowstyle{\bfseries}}   

\definecolor{ansi-red-intense}{HTML}{B22B31}
\definecolor{ansi-blue-intense}{HTML}{0065CA}
\usepackage{fancyvrb} 
\DefineVerbatimEnvironment{Highlighting}{Verbatim}{commandchars=\\\{\}}
    \makeatletter
        \newbox\Wrappedcontinuationbox 
        \newbox\Wrappedvisiblespacebox 
        \newcommand*\Wrappedvisiblespace {\textcolor{red}{\textvisiblespace}} 
        \newcommand*\Wrappedcontinuationsymbol {\textcolor{red}{\llap{\tiny$\m@th\hookrightarrow$}}} 
        \newcommand*\Wrappedcontinuationindent {3ex } 
        \newcommand*\Wrappedafterbreak {\kern\Wrappedcontinuationindent\copy\Wrappedcontinuationbox} 
        \newcommand*\Wrappedbreaksatspecials {%
            \def\PYGZus{\discretionary{\char`\_}{\Wrappedafterbreak}{\char`\_}}%
            \def\PYGZob{\discretionary{}{\Wrappedafterbreak\char`\{}{\char`\{}}%
            \def\PYGZcb{\discretionary{\char`\}}{\Wrappedafterbreak}{\char`\}}}%
            \def\PYGZca{\discretionary{\char`\^}{\Wrappedafterbreak}{\char`\^}}%
            \def\PYGZam{\discretionary{\char`\&}{\Wrappedafterbreak}{\char`\&}}%
            \def\PYGZlt{\discretionary{}{\Wrappedafterbreak\char`\<}{\char`\<}}%
            \def\PYGZgt{\discretionary{\char`\>}{\Wrappedafterbreak}{\char`\>}}%
            \def\PYGZsh{\discretionary{}{\Wrappedafterbreak\char`\#}{\char`\#}}%
            \def\PYGZpc{\discretionary{}{\Wrappedafterbreak\char`\%}{\char`\%}}%
            \def\PYGZdl{\discretionary{}{\Wrappedafterbreak\char`\$}{\char`\$}}%
            \def\PYGZhy{\discretionary{\char`\-}{\Wrappedafterbreak}{\char`\-}}%
            \def\PYGZsq{\discretionary{}{\Wrappedafterbreak\textquotesingle}{\textquotesingle}}%
            \def\PYGZdq{\discretionary{}{\Wrappedafterbreak\char`\"}{\char`\"}}%
            \def\PYGZti{\discretionary{\char`\~}{\Wrappedafterbreak}{\char`\~}}%
        } 
        \newcommand*\Wrappedbreaksatpunct {%
            \lccode`\~`\.\lowercase{\def~}{\discretionary{\hbox{\char`\.}}{\Wrappedafterbreak}{\hbox{\char`\.}}}%
            \lccode`\~`\,\lowercase{\def~}{\discretionary{\hbox{\char`\,}}{\Wrappedafterbreak}{\hbox{\char`\,}}}%
            \lccode`\~`\;\lowercase{\def~}{\discretionary{\hbox{\char`\;}}{\Wrappedafterbreak}{\hbox{\char`\;}}}%
            \lccode`\~`\:\lowercase{\def~}{\discretionary{\hbox{\char`\:}}{\Wrappedafterbreak}{\hbox{\char`\:}}}%
            \lccode`\~`\?\lowercase{\def~}{\discretionary{\hbox{\char`\?}}{\Wrappedafterbreak}{\hbox{\char`\?}}}%
            \lccode`\~`\!\lowercase{\def~}{\discretionary{\hbox{\char`\!}}{\Wrappedafterbreak}{\hbox{\char`\!}}}%
            \lccode`\~`\/\lowercase{\def~}{\discretionary{\hbox{\char`\/}}{\Wrappedafterbreak}{\hbox{\char`\/}}}%
            \catcode`\.\active
            \catcode`\,\active 
            \catcode`\;\active
            \catcode`\:\active
            \catcode`\?\active
            \catcode`\!\active
            \catcode`\/\active 
            \lccode`\~`\~ 	
        }
    \makeatother

    \let\OriginalVerbatim=\Verbatim
    \makeatletter
    \renewcommand{\Verbatim}[1][1]{%
        \sbox\Wrappedcontinuationbox {\Wrappedcontinuationsymbol}%
        \sbox\Wrappedvisiblespacebox {\FV@SetupFont\Wrappedvisiblespace}%
        \def\FancyVerbFormatLine ##1{\hsize\linewidth
            \vtop{\raggedright\hyphenpenalty\z@\exhyphenpenalty\z@
                \doublehyphendemerits\z@\finalhyphendemerits\z@
                \strut ##1\strut}%
        }%
        
        \Wrappedbreaksatspecials
        \OriginalVerbatim[#1,codes*=\Wrappedbreaksatpunct]%
    }
    \makeatother
\usepackage{tikz}

\definecolor{colchristin}{rgb}{1, 0.3, 0.3}
\definecolor{colbach}{rgb}{0.3, 0.3, 1}
\definecolor{coljoerg}{rgb}{0.3, 1, 0.3}

\usepackage{color, colortbl}
\usepackage{multirow}
\usepackage[labelfont=bf]{caption}
\usepackage{subcaption}
\definecolor{Gray}{gray}{0.9}
\sethlcolor{Gray}
\def\BibTeX{{\rm B\kern-.05em{\sc i\kern-.025em b}\kern-.08em
    T\kern-.1667em\lower.7ex\hbox{E}\kern-.125emX}}
\usepackage{graphicx}
\newcolumntype{P}[1]{>{\centering\arraybackslash}p{#1}}
\begin{document}
\title{From Black Boxes to Conversations: Incorporating XAI in a Conversational Agent}
\titlerunning{Incorporating XAI in a Conversational Agent}
%
\author{Van Bach Nguyen\inst{1,2}\orcidID{0000-0002-4576-9302} \and
Jörg Schlötterer\inst{1,2,3}\orcidID{0000-0002-3678-0390} \and
Christin Seifert\inst{1,2}\orcidID{0000-0002-6776-3868}}
\authorrunning{Van Bach Nguyen et al.}
%
\institute{University of Marburg, Germany \and
University of Duisburg-Essen, Germany\\
\and
University of Mannheim, Germany\\
\email{
nvbach92@gmail.com\\
\{joerg.schloetterer, christin.seifert\}@uni-marburg.de
}
}

\maketitle              
\begin{abstract}
The goal of Explainable AI (XAI) is to design methods to provide insights into the reasoning process of black-box models, such as deep neural networks, in order to explain them to humans. 
Social science research states that such explanations should be conversational, similar to human-to-human explanations. 
In this work, we show how to incorporate XAI in a conversational agent, using a standard design for the agent comprising natural language understanding and generation components.
We build upon an XAI question bank, which we extend by quality-controlled paraphrases, to understand the user's information needs. We further systematically survey the literature for suitable explanation methods that provide the information to answer those questions, and present a comprehensive list of suggestions. 
Our work is the first step towards truly natural conversations about machine learning models with an explanation agent. The comprehensive list of XAI questions and the corresponding explanation methods may support other researchers in providing the necessary information to address users' demands. 
To facilitate future work, we release our source code and data.\footnote{ \url{https://github.com/bach1292/XAGENT/}\label{fn:source}}

\end{abstract}
\section{Introduction}
\label{sec:intro}

Deep neural networks~\cite{Goodfellow2016} are the de-facto standard in machine learning-based artificial intelligence (AI) systems, having shown to outperform humans on many tasks (e.g.,~\cite{McKinney2020AI}).
However, there is a trade-off between predictive performance and interpretability -- while modern AI systems are highly powerful, they are also black-boxes by nature and their reasoning process is not interpretable anymore~\cite{guidotti_survey_2018}.
This tradeoff is addressed by the young, but rapidly growing research field of XAI~\cite{barredo_arrieta_explainable_2020}.

While much progress has been made in XAI in terms of algorithmic development in recent years, this progress is not accessible to all stakeholders, in particular not to end users~\cite{liao_questioning_2020}.
Miller~\cite{miller_explanation_2019} argues that human explanations are social conversations and, in order for XAI to be truly interpretable and valuable for all stakeholders, XAI needs to be conversational as well.
These conversations are not restricted to natural language, but explanations could also be presented in a visual manner.
The maxim is to provide explanations in a conversational manner, with interactions based on questions and answers.
\begin{figure}[t]
\scriptsize
{\ttfamily
\textcolor{ansi-blue-intense}{\textbf{X-Agent: }} I recorded the profile: [Age 39, Workclass: State-gov, Education-num: 15, Marital-status: Separated, Occupation: Other-service, Relationship: Not-in-family, Race: White, Sex: Male, Capital-gain: 0, Capital-loss: 0, Hour-per-week: 20.0, Native-country: England].\\With this profile, the income will be <=50K.\\[2pt]
\textcolor{ansi-red-intense}{\textbf{User:}} Give me the reason for this prediction!\\
\textcolor{ansi-blue-intense}{\textbf{X-Agent: }}\\
\includegraphics[width=\linewidth]{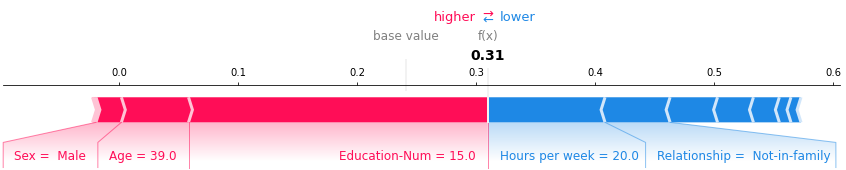}
The above graph shows important features for this prediction. The red features increase the income, while the blue features decrease it.\\[2pt]
\textcolor{ansi-red-intense}{\textbf{User:}} Why is this profile predicted <=50K instead of >50K?\\
\textcolor{ansi-blue-intense}{\textbf{X-Agent: }} There are multiple reasons for this result:\\Age is too low and Workclass is not suitable.\\If Age increases to at least 66.3 and Workclass changes to Self-emp-inc, this person's predicted income will be more than 50K. \\[2pt]
\textcolor{ansi-red-intense}{\textbf{User:}} That's hard, how could I change only Occupation to get >50K prediction?\\
\textcolor{ansi-blue-intense}{\textbf{X-Agent: }} The Occupation should be changed to Tech-support to get an income of more than 50K.
}
\caption{XAI user questions are matched to reference questions. Answers are provided by a mapping of reference question to XAI methods along with additional explanations. The X-Agent explains the model's reasoning by feature importance scores of SHAP~\cite{lundberg_unified_2017} and an explanation how to interpret them.}
    \label{fig:result:example_adult}
\end{figure}
This maxim is explicitly expressed by practitioners~\cite{slack2022desiderata}.
While the need for conversational XAI has been recognized, fulfilling it remains an open challenge with solutions either limited to small, specific data sets and models~\cite{kuzba_conversational_2021} or mere collections of requirements for conversational XAI~\cite{werner_explainable_2020}. 
In this work, we develop methods to leverage a standard conversational agent architecture to conversational XAI. 
Building upon well-established conversational agent techniques allows us to focus on XAI-specific requirements, in order to cover a broad range of user questions, types of data, and types of models.

Our target group is users without machine learning knowledge who interact with the system. Following the taxonomy by Tomset et al.~\cite{tomsett2018interpretable}, users can be executors or operators, e.g., doctors making decisions based on the system's advice, or lenders using systems to assess applicant profiles. Figure~\ref{fig:result:example_adult} shows an example of a conversation between our prototype agent and a user asking to explain a prediction of a Random Forest model on the Adult dataset\footnote{\url{https://archive.ics.uci.edu/ml/datasets/adult/}\label{fn:adult_data}} (details on each individual interaction step, as well as a conversational scenario about a Convolutional Neural Network model on image data can be found in the Appendix, Section~\ref{sec:scenarios:cnn-on-mnist}).

\begin{figure*}[h]
    \centering
    \includegraphics[width=\textwidth]{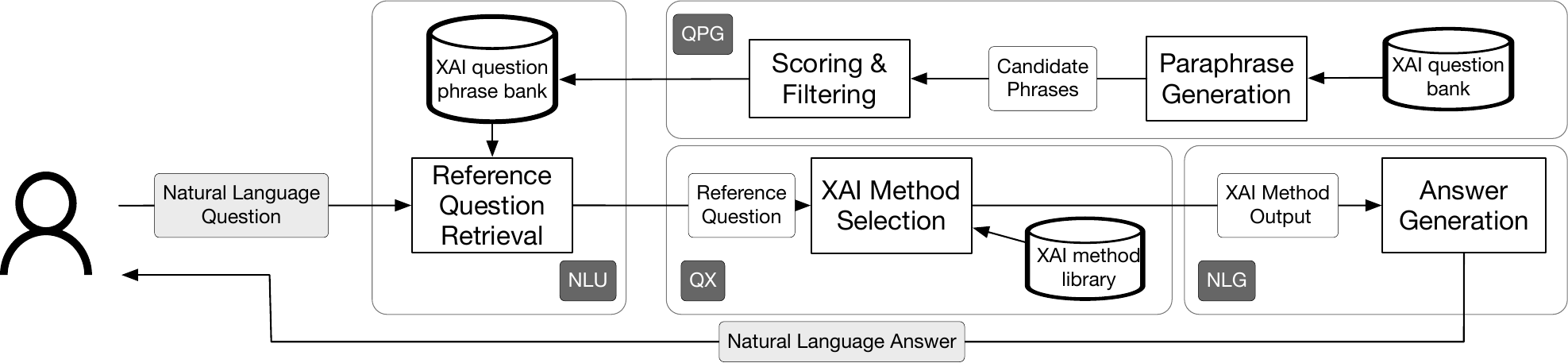}
    \caption{Incorporating XAI in conversational agents: 1) Question-Phrase-Generation (QPG) uses a paraphrase generation model on the questions from the XAI question bank~\cite{liao_questioning_2020}. The generated candidates are scored by multiple annotators and ranked, resulting in the XAI question phrase bank. 2) In the Natural Language Understanding (NLU) component, the reference question for a user question is retrieved from the phrase bank. 3) The intent of the reference question defines the XAI method to be applied to the model in the Question-XAI method mapping component (QX). 4) A natural language generation (NLG) component converts the output of the XAI method (e.g., a table, graph, or number) with an answer in natural language. Omitted for overview: data sets are loaded and machine learning models are trained dynamically on user requests. }
    \label{fig:approach:overview}
\end{figure*}

We propose a systematic approach to enable XAI in a standard conversational agent architecture~\cite{bobrow1977}  (see Figure~\ref{fig:approach:overview} for an overview, full details in Section~\ref{sec:sysoverview}).
First, we construct a question phrase bank data set based on the question bank of Liao et al.~\cite{liao_questioning_2020} to enable the agent to understand XAI questions. 
Second, based on a systematic literature survey of explanation methods, we establish a mapping between user intents, represented as reference questions in the question phrase bank, and XAI methods to answer user questions. 
Finally, a template-based natural language generation component creates the answers using the information from the selected XAI methods. 
Specifically, our contributions are:
\begin{itemize}
\item 
  We present a systematic overview of methods to answer the information need implied in these questions and categorise questions by identifying which subsets require an XAI method for answering.
   \item We create a publicly available XAI question phrase bank for training and evaluating the natural language understanding component of the conversational agent based on the XAI question bank of Liao et al.~\cite{liao_questioning_2020}.\footref{fn:source}
  \item 
  We incorporate XAI in a standard conversational agent framework and present a prototype that can communicate about the internals of the machine learning model in natural language.\footref{fn:source}

\end{itemize}

\section{Related Work}
\label{sec:rel-work}

XAI is a highly active research field from the algorithmic perspective (see \cite{ali2023explainable} for a recent survey).
While the four key XAI surveys~\cite{guidotti_survey_2018,barredo_arrieta_explainable_2020,gilpin_explaining_2019,adadi_peeking_2018} focus on different perspectives, a common pattern in their taxonomies is the distinction between intrinsically interpretable models, i.e., the model itself constitutes the explanation, and posthoc explanations, i.e., methods explaining black-box models by, e.g., linear models approximating local decision boundaries~\cite{ribeiro_why_2016}.
Posthoc explanations can be further classified as model-specific, applicable only to certain types of models or model-agnostic, applicable to any machine learning model.
We focus on model-agnostic posthoc explanations to be able to capture a broad range of machine learning tasks and models in the conversational agent.
We derive posthoc explanation methods, which are suitable to answer specific XAI questions from the aforementioned surveys, and additionally from the suggested methods for each group of XAI questions in~\cite{liao_questioning_2020,liao_human-centered_2022}.

Recent research in conversational agents has gradually shifted to end-to-end approaches, based on fully data-driven systems~\cite{Jianfeng2018Neutral}.  However, the abundance of data is an ultimate prerequisite of such systems, and large data sets are not available for XAI yet.
In contrast, the Genial Understander System (GUS)~\cite{bobrow1977}, which underlies most modern commercial digital assistants~\cite{jurafsky2000speech}, requires only a small amount of data to build a conversational agent.
Therefore, we focus on enabling XAI in this framework.
Since public data sets are limited to a few domains~\cite{rastogi_towards_2020} and not available for XAI, we create a question phrase corpus to fill this gap.
Research on Conversational XAI is still in its infancy, and agents are strongly limited in scope. Werner~\cite{werner_explainable_2020} presents a work-in-progress prototype of a rule-based XAI chatbot to iteratively elaborate requirements, accompanied by findings from literature and user studies.
His prototype is limited to the classification of tabular data and a small set of pre-defined questions.
Kuźba and Biecek's~\cite{kuzba_conversational_2021} goal is to collect the needs of users interacting with a conversational XAI agent, i.e., questions a human would ask.
Their prototype is limited to a Random Forest, applied to the Titanic Dataset, and explanations addressing a few types of XAI questions, since the primary goal is the collection of interaction data.
Contrary to the data-driven approach of Kuźba and Biecek, Liao et al.~\cite{liao_questioning_2020} construct an XAI question bank from literature review, expert reviews by XAI practitioners, and around 1000 minutes of interviews with 20 user experience and design practitioners working in multiple domains.
This question bank contains 73 XAI questions in 10 categories (see Table~\ref{tab:approach:xai-question-and-methods}, columns 2 and 3) and serves as the basis for our question phrase bank.
While the aforementioned work collects users' needs and questions to (conversational) XAI, our goal is to automatically provide answers to the identified user questions.
Developed at the same time, the conversational XAI agent by Slack et al.~\cite{slack_talktomodel_2022} is most similar to our work, in particular in using language models to identify users' intents and template-based answer generation.
However, their focus is on open-ended natural language dialogues and hence their agent is limited to tabular data, whereas we focus on a broad coverage of XAI questions, including a multitude of data types and models. 

\section{System overview}
\label{sec:sysoverview}
A standard conversational agent architecture~\cite{bobrow1977} is generally composed of the following components: i) Natural Language Understanding (NLU) to, e.g., determine a user's intent(s), ii) a dialogue state tracker (DST) to maintain the current state and history of the dialogue, iii) a dialogue policy, deciding the system's next step, and iv) Natural Language Generation (NLG) to generate the system's output. We integrate XAI into this architecture and assume that an XAI question contains all relevant information to select a proper explanation method as a response (the extension to incomplete information is subject to future work). Hence, we omit the DST.
This section outlines our approach to incorporate XAI into the remaining components.

The general architecture of our conversational XAI agent is depicted in Figure~\ref{fig:approach:overview}. 
The NLU component is responsible for identifying the user's actual intent from a wide range of XAI utterances.
To cope with this variety of utterances, we expand the XAI question bank~\cite{liao_questioning_2020} into an XAI question phrase bank.
This phrase bank constitutes a training data set to identify user intents from a wide range of XAI utterances.
We construct the extended question phrase bank from an initial set of XAI questions, paraphrase generation, and scoring. We describe the construction (QPG) in full detail in \cref{sec:question-phrase-bank}. 
Upon a user question, the NLU component retrieves the corresponding reference question from the question phrase bank. We explain this retrieval in Section~\ref{ssec:approach:nlu}. 
The intent of the reference question determines the XAI method to be applied to the model in the Question-XAI method mapping, which is our integration of XAI  to the dialogue policy component (QX, see Section~\ref{ssec:approach:qx}). 
The natural language generation (NLG) component \cite{reiter1997building, gatt2018survey} enriches the output  of the XAI method, e.g., SHAP, by natural language to form the final answer, e.g., explaining SHAP's output graph  (see Section~\ref{sec:approach:nlg}).

\section{Understanding Questions} 

In this section, we first describe the construction of the question phrase bank, which helps the agent understand a broad set of XAI questions, and then describe how reference questions can be retrieved from this question phrase bank to implement the NLU component.

\subsection{Question Phrase Bank (QPG)}
\label{sec:question-phrase-bank}
Training the NLU component of a conversational XAI agent requires conversational data about XAI. 
However, such data does not exist. Therefore, we introduce an XAI question phrase bank as a data set to train and evaluate the NLU component. The phrase bank represents a broad variety of utterances of possible user questions to XAI systems and is publicly available.\footref{fn:source}

To create the phrase bank, we use a language model to generate paraphrase candidates for each reference question in the XAI question bank of Liao et al.~\cite{liao_questioning_2020}. Our goal is to capture a high variance of utterances in the phrase bank while retaining semantics of the reference questions in the initial question bank. The paraphrase candidates were manually scored for similarity by multiple annotators before we filter them by their scores (Figure~\ref{fig:approach:overview}, QPG component, top right).

We use GPT-3~\cite{brown_language_2020} and few-shot learning  to generate paraphrase candidates of XAI questions\footnote{We use the Open AI API:  {\url{https://openai.com/api/}}} (for details and an example see Appendix~\ref{asec:gpt-3-paraphrases}).
During the paraphrase generation process, we discard any generated text that does not clearly constitute a paraphrased question (e.g., answers to  questions, incomplete sentences). 
We generate at least 2 paraphrases per question, 4.2 on average, 20 at maximum, and 310 pairs of (question, paraphrase candidate) in total.

    

To assess the quality of the paraphrase candidates, we annotated all generated pairs manually by human-perceived similarity.  We extended our paraphrase candidate set by 59 negative pairs, i.e., we sampled paraphrases from different, non-matching questions at random. Thus, our data set for annotation comprises 369 phrase pairs. 
Annotators were first introduced to the task, shown a simple example, and then asked to assess the similarity of phrase pairs on a 6-point Likert scale (1: very different, 6: very similar)~\cite{amidei_use_2019}. We chose a scale with an even number of items to force respondents to select between either similar or different because a neutral element ``halfway similar and halfway different`` is neither meaningful nor can it be assessed.
Seven participants, consisting of five males and two females, with a Master's or Ph.D. degree in computer science and a background in machine learning, took part in the annotation process.

We randomly assigned participants to one of 3 groups, one participant was assigned to all groups. Each group annotated approx. 125 phrase pairs, resulting in each pair having at least 2 annotations. 
In the final annotation scores, most of the paraphrases generated by GPT-3 have a score $\geq 4$, while most of negatives pairs have a score $<4$ supporting the high quality of GPT-3 paraphrases as well as human annotation. Further details on the scores can be found in the Appendix, Section~\ref{sec:append:annotation}.

For our final XAI question phrase bank, we select all pairs of paraphrases with an average annotation score $\geq 4$ (Likert score of 4 means \textit{more similar than different}). Each paraphrase is linked to its reference question, and the reference question is a paraphrase of itself. The task of the NLU component is then to identify the reference question for a user question, based on the known paraphrases of the reference question. 

\subsection{Reference Question Retrieval (NLU)}
\label{ssec:approach:nlu}
We preprocess a given user question to a standard format. We use a placeholder $<$feature$>$ to substitute all feature names from the data set. Similarly, labels (classes) in user questions are replaced by the placeholder $<$class$>$. 
For example, on the Adult data set\footref{fn:adult_data} (Figure~\ref{fig:result:example_adult}), the question \textit{How could I change only Occupation to get \textgreater50K?} is transformed to \textit{How could I change only $<$feature$>$ to get $<$class$>$?}, in which, the \textit{Occupation} is a feature and \textit{\textgreater50K} is a class in the Adult data set.
We assume that feature and class names in user questions match those in the data set (i.e., no typos or synonyms).
We formulate the matching of a user question to a reference question as a multi-class classification problem with class labels corresponding to reference questions in the XAI question phrase bank (see~\cref{sec:question-phrase-bank}).
First, we generate sentence embeddings of the pre-processed user and reference questions with SimCSE~\cite{gao_simcse_2021} and RoBERTa-large~\cite{liu_roberta_2019}.  
We then train a feedforward network with 1 hidden layer and ReLU activation on the sentence embeddings to classify user questions into one of the reference questions in the question phrase bank.
The output of this step is a reference question that reflects the intent of the user question. From the classifier's output, we select the reference question with the highest probability.
If the probability is lower than a predefined threshold $\theta$ (no match), we consider the question as an (yet) unknown variation (paraphrase) of a reference, save it for later, and ask the user for an alternative phrasing of the question. We set $\theta=0.5$ in our experiments. We provide an evaluation of this matching approach and a comparison to other approaches in section~\ref{sec:eval}.

In summary, given a user question, we substitute all feature names and class names in the question with placeholders. Then, we use a pretrained model to match the preprocessed question to one of the reference questions in the question phrase bank, which we created using GPT-3 and human annotation. 

\section{Retrieving Answer Information}
\label{ssec:approach:qx}
After matching user input to its corresponding reference question, the next step is to obtain the relevant information to provide an answer. Previous work~\cite{liao_questioning_2020,liao_human-centered_2022} suggested some XAI methods as responses for question groups, but it remains unclear how to select the appropriate method for each specific question, i.e., how to design a simplified dialogue policy in a conversational XAI agent (QX component in~\cref{fig:approach:overview}). 

{\scriptsize
\begin{longtable}{+c^l^p{3.6cm}^c^p{2.5cm}^L{2.5cm}}
\caption{Overview of XAI questions~\cite{liao_questioning_2020} with reference questions, number of paraphrases (\textbf{Phr.}), whether the question requires an XAI method (\hl{highlighted rows}, see \cref{ssec:approach:qx:xai-questions}), and (optional) sources of information~(\cref{ssec:answer:nonXAI}). ``n.a.'': no method matches the selection criteria; \textbf{bold} indicates the final selected methods. Options are not always  available, limited to certain types of data/models or provide only partial information.}
\label{tab:approach:xai-question-and-methods} \\
\toprule\tabhead
ID& Category  & Reference Question & Phr. & Method & Option \\\otoprule
1 & How & How are the parameters set? & 6 & Model Generation & ModelCards~\cite{Mitchel2019_fatml_modelcards}\\
\rowcolor{Gray}
2 & How & How does feature f impact its predictions? & 4  & \textbf{SHAP}~\cite{lundberg_unified_2017}, (LIME~\cite{ribeiro_why_2016}) & \\
\rowcolor{Gray}
3 & How & How does it weigh different features? & 7  & \textbf{SHAP}~\cite{lundberg_unified_2017}, (LIME~\cite{ribeiro_why_2016}) & \\
\rowcolor{Gray}
4 & How & How does the system make predictions? & 4  & n.a. & ProtoTree~\cite{nauta2021neural}, ProtoPNet~\cite{chen2019looks}, ModelExtraction~\cite{bastani2017interpretability}\\
\rowcolor{Gray}5 & How & Is feature X used or not used for the predictions? & 4  & \textbf{SHAP}~\cite{lundberg_unified_2017}, (LIME~\cite{ribeiro_why_2016}) & \\
\rowcolor{Gray}6 & How & What are the top features it uses? & 4  & \textbf{SHAP}~\cite{lundberg_unified_2017}, (LIME~\cite{ribeiro_why_2016}) & \\
\rowcolor{Gray}7 & How & What are the top rules it uses? & 4  & n.a. & \\
\rowcolor{Gray}8 & How & What features does the system consider? &   & \textbf{SHAP}~\cite{lundberg_unified_2017}, (LIME~\cite{ribeiro_why_2016}) & \\
\rowcolor{Gray}9 & How & What is the system's overall logic? & 4  & n.a. & ProtoTree~\cite{nauta2021neural}, ProtoPNet~\cite{chen2019looks}, ModelExtraction~\cite{bastani2017interpretability}\\
10 & How & What kind of algorithm is used? & 4 & Model Generation & ModelCards~\cite{Mitchel2019_fatml_modelcards}\\
\rowcolor{Gray}11 & How & What rules does it use? & 4  & n.a. & ProtoTree~\cite{nauta2021neural}\\
\midrule
\rowcolor{Gray}12, 13 & How to be & How should this instance/feature change to get a different prediction? & 7  & \parbox[t]{2.5cm}{\textbf{DICE}~\cite{mothilal_explaining_2020}, \\ \textbf{CFProto}~\cite{van_looveren_interpretable_2020} for  12} & \\
\rowcolor{Gray}14 & How to be & What kind of instance gets a different prediction? & 4  & \textbf{DICE}~\cite{mothilal_explaining_2020}, \textbf{CFProto}~\cite{van_looveren_interpretable_2020} & \\
\midrule
\rowcolor{Gray}15, 16 & How to still & What are the necessary features present/ absent to guarantee this prediction? & 6  & n.a. & SHAP~\cite{lundberg_unified_2017}, (LIME~\cite{ribeiro_why_2016})\\
\rowcolor{Gray}17, 18 & How to still & What is the highest/lowest feature one can have to still get the same prediction? & 12  & \textbf{Anchors}~\cite{ribeiro_anchors_2018} & \\
\rowcolor{Gray}19 & How to still & What is the scope of change permitted to still get the same prediction? & 4  & \textbf{Anchors}~\cite{ribeiro_anchors_2018} & \\
\rowcolor{Gray}20 & How to still & What kind of instance gets this prediction? & 4  & \textbf{Anchors}~\cite{ribeiro_anchors_2018} & \\
\midrule
21 & Input & How much data like this is the system trained on? & 4 & Model Generation & ModelCards~\cite{Mitchel2019_fatml_modelcards}, DataSheets~\cite{Gebru2021_commacm_datasheets}\\
22, 23 & Input & How were the ground-truth/labels produced? & 8 & Data Generation & DataSheets~\cite{Gebru2021_commacm_datasheets}\\
24, 25 & Input & What are the biases/limitations of the data? & 9 & Data Generation & DataSheets~\cite{Gebru2021_commacm_datasheets}\\
26 & Input & What data is the system not using? & 5 & Model Generation & ModelCards~\cite{Mitchel2019_fatml_modelcards}\\
27 & Input & What is the sample size? & 3 & Model Generation & ModelCards~\cite{Mitchel2019_fatml_modelcards}, DataSheets~\cite{Gebru2021_commacm_datasheets}\\
28 & Input & What is the source of the data? & 3 & Data Generation & ModelCards~\cite{Mitchel2019_fatml_modelcards}, DataSheets~\cite{Gebru2021_commacm_datasheets}\\
29 & Input & What kind of data does the system learn from? & 5 & Model Generation & ModelCards~\cite{Mitchel2019_fatml_modelcards}, DataSheets~\cite{Gebru2021_commacm_datasheets}\\
\midrule
30 & Output & How can I best utilize the output of the system? & 4 & Model Generation & ModelCards~\cite{Mitchel2019_fatml_modelcards}, DataSheets~\cite{Gebru2021_commacm_datasheets}\\
31 & Output & How is the output used for other system component(s)? & 4 & System Context & \\
32 & Output & What does the system output mean? & 4 & Data/Model Generation & ModelCards~\cite{Mitchel2019_fatml_modelcards}, DataSheets~\cite{Gebru2021_commacm_datasheets}\\
33 & Output & What is the scope of the system's capability? Can it do [A]? & 4 & Model Generation & ModelCards~\cite{Mitchel2019_fatml_modelcards}\\
34 & Output & What kind of output does the system give? & 3 & Data/Model Generation & ModelCards~\cite{Mitchel2019_fatml_modelcards}, DataSheets~\cite{Gebru2021_commacm_datasheets}\\
\midrule
35--37 & Performance & How accurate/precise/reliable are the predictions? & 12 & Model Generation & ModelCards~\cite{Mitchel2019_fatml_modelcards}\\
38 & Performance & How often does the system make mistakes? & 4 & Model Generation & ModelCards~\cite{Mitchel2019_fatml_modelcards}\\
39, 40 & Performance & In what situations is the system likely to be correct/incorrect? & 8 & Model Generation & ModelCards~\cite{Mitchel2019_fatml_modelcards}\\
41 & Performance & Is the system's performance good enough for [A]? & 2 & System Context & \\
42 & Performance & What are the limitations of the system? & 2 & Model Generation & ModelCards~\cite{Mitchel2019_fatml_modelcards}\\
43 & Performance & What kind of mistakes is the system likely to make? & 5 & Model Generation & ModelCards~\cite{Mitchel2019_fatml_modelcards}\\
\midrule
44 & What if & What would the system predict for [a different instance]? & 2 & Prediction & \\
45, 46 & What if & What would the system predict if feature(s) f of this instance change(s) to f’? & 8 & Prediction & \\
\midrule
\rowcolor{Gray}47--48 & Why & Why/How is this instance given this prediction? & 20  & \textbf{SHAP}~\cite{lundberg_unified_2017}, (LIME~\cite{ribeiro_why_2016}) & \\
\rowcolor{Gray}49 & Why & What features of this instance lead to the system's prediction? & 15  & \textbf{SHAP}~\cite{lundberg_unified_2017}, (LIME~\cite{ribeiro_why_2016}) & \\
\rowcolor{Gray}50 & Why & Why are instance A and instance B given the same prediction? & 4  & \textbf{SHAP}~\cite{lundberg_unified_2017}, (LIME~\cite{ribeiro_why_2016}) & \\
\midrule
\rowcolor{Gray}51 & Why not & How is this instance not predicted A? & 4  & \textbf{DICE}~\cite{mothilal_explaining_2020}, \textbf{CFProto}~\cite{van_looveren_interpretable_2020} & \\
\rowcolor{Gray}52 & Why not & Why are instances A and B given different predictions? & 8  & \textbf{DICE}~\cite{mothilal_explaining_2020}, \textbf{CFProto}~\cite{van_looveren_interpretable_2020} & \\
\rowcolor{Gray}53 & Why not & Why is this instance predicted P instead of Q? & 11  & \textbf{DICE}~\cite{mothilal_explaining_2020}, \textbf{CFProto}~\cite{van_looveren_interpretable_2020} & \\
\midrule
54 & Others & How to improve the system? & 4 & External Knowledge & \\
55 & Others & What are the results of other people using the system? & 4 & External Validation & \\
56 & Others & What does [ML terminology] mean? & 2 & External Knowledge & \\
57-- 67 & Others & How/What/Why will the system adapt/change/drift/improve over time? & 35 & External Validation & \\
\rowcolor{Gray}68--70 & Others & Why NOT using this data/feature/rule? & 8  & n.a. & \\
\rowcolor{Gray}71--73 & Others & Why using this data/feature/rule? & 14  & n.a. & \\\bottomrule
\end{longtable}
}

We analyzed all 73 reference questions in the XAI question phrase bank for their implied information need and identified methods to retrieve this information. 
Table~\ref{tab:approach:xai-question-and-methods} presents an overview of all questions. Specifically, we identified the questions that require an XAI method for extracting the answer information (highlighted rows), and not only require to access stored values, e.g. the size of the training data. 
Following the general definition of XAI by Arrieta et al.~\cite{barredo_arrieta_explainable_2020}, we define an XAI method\footnote{
By defining XAI methods, our goal is to distinguish between approaches that rely on models' internal reasoning and those that only involve simple actions such as retrieving information or making predictions using the model.} as a method that produces details or reasons to make the AI's functioning clear or easy to understand. 
That is, an XAI method must have access to a model's internal reasoning or to a proxy that reveals this reasoning at least to some extent.
For instance, the question \textit{Why is this instance predicted  P instead of Q?} requires a counterfactual explanation, identifying feature sets that -- if changed -- would change the model's prediction from P to the specified counterfactual class Q. On the other hand, the question \textit{What would the system predict if feature [..] of this instance changed to f'}, with f' given as specific feature values, just requires to create a new test instance with the specific feature set and apply the trained (black-box) model on this new instance. 

In the following, we first discuss the information needed for the 50 questions that do not require an XAI method, and outline how the information for the answer can be obtained. Second, we discuss the 23 XAI questions
and present our criteria for analysing existing XAI methods and our final mapping from reference question to XAI-method for extracting the answer information. 

\subsection{Non XAI Specific Questions}
\label{ssec:answer:nonXAI}
Analysing the 50 questions in this category, we identified six subcategories w.r.t. information need. The categories are indicated in column ``Method'' in  Table~\ref{tab:approach:xai-question-and-methods}.
Questions in the \textbf{data generation} subcategory require information about the data or the data generation process. They can either be directly answered by querying data set statistics or accessing an accompanying data sheet for the data set~\cite{Gebru2021_commacm_datasheets} if available. Examples of such questions are \textit{What is the sample size?} or \textit{How were the ground-truth labels produced?} 
Questions in the \textbf{model generation} subcategory can be answered by retrieving easily accessible information about the underlying machine learning model, such as hyperparameters set during training or evaluation results. If the model is equipped with a ModelCard~\cite{Mitchel2019_fatml_modelcards}, the information can be obtained from the latter. Example questions for this category are \textit{How often does the system make mistakes?} and \textit{What kind of algorithm is used?}
ModelCards also contain information about biases, scope, and limitations of the machine learning model, thus containing information for other questions, such as \textit{What is the scope of the systems ability?}. 
Questions in the \textbf{prediction} subcategory, such as \textit{What would the system predict for [a different instance]?}, can simply be answered by applying the model on a newly generated test instance. 
Some questions require \textbf{external knowledge}, either additional information from humans or an additional external knowledge base and an information retrieval approach, to access the information. For example, the question \textit{How to improve the system?} requires domain knowledge on model optimization, a judgment of model performance in comparison with similar approaches, and, for instance, an estimate of whether additional training data are likely to improve performance. For the question \textit{What does [ML terminology] mean?}, we envision a  knowledge base or lexicon with Machine Learning terms that can be queried for the information. 
Questions in the \textbf{external validation} subcategory, such as \textit{How will the system improve over time}, require additional evaluation during the system's lifetime and/or information from similar systems. The simplest and easily accessible information would be a binary indicator of whether the system is capable of online learning at all, but more  details require elaborate experiments. 
\textbf{System context}  questions ask for information about the integration of the machine learning model with other system components and their embedding in the application. E.g., the question \textit{How is the output used for other system component(s)?} depends on the actual system deployment, and \textit{Is the system good enough for [A]?} requires knowledge about the application context and its requirements.

\subsection{XAI Questions}
\label{ssec:approach:qx:xai-questions}

To identify suitable candidates able to provide the information for answering the 23 XAI questions remaining
(highlighted rows in Table~\ref{tab:approach:xai-question-and-methods}), we conducted a literature survey. 
Our sources were the four key XAI surveys~\cite{guidotti_survey_2018,barredo_arrieta_explainable_2020,gilpin_explaining_2019,adadi_peeking_2018}, the methods referred to in the initial XAI question bank and the follow-up work by the same authors~\cite{liao_questioning_2020,liao_human-centered_2022}. 
To further filter the candidate set for XAI methods to incorporate in conversational agents, we defined the following four criteria for our analysis, the latter three based on the categorization scheme of a recent survey~\cite{nauta_anecdotal_2022}, which identified 312 original papers, presenting a novel XAI technique in one of the major ML/AI conferences since 2014 until 2020:

\textbf{Source Code:} The paper should be accompanied by easy-to-use, publicly accessible source code in order to integrate the method into the conversational agent in a reasonable amount of time. This criterion significantly reduced the number of methods.

\textbf{Model:} We assessed the applicability of the XAI method to different types of models and favored approaches that are model agnostic~\cite{barredo_arrieta_explainable_2020} or at least can be applied to multiple types of models. 

\textbf{Data:} We assessed the applicability of the XAI method to different types of input data. To enable efficient implementation of components that process the XAI method's raw output to generate a user-friendly natural language answer, possibly accompanied by visualizations, we favor methods that can be applied to multiple types of input data.

\textbf{Problem:} Explanation methods may be restricted to particular machine learning tasks, e.g., regression. We account for this restriction by including the category ``type of problem'' in our analysis. In this paper, we focus on explaining models for supervised machine learning, more specifically classification tasks. 

We provide the detailed overview in the Appendix, Table~\ref{tab:apdx:xai-method-and-questions} and focus on the finally selected methods in this section. Table~\ref{tab:approach:xai-question-and-methods} shows the selected methods (in \textbf{bold}) and their corresponding questions for which they provide the necessary information. 
We briefly describe the selected methods in the following:


\noindent SHAP~\cite{lundberg_unified_2017} is a widely used method to quantify the importance of features for the prediction of a single instance. SHAP follows a game-theoretic approach to identify the contribution of each feature (player) in an additive setting. Quantitative feature importance values are relevant for answering questions w.r.t. feature contribution on the prediction (questions 2, 3, 5, 8, 48) as well as for top features (question 6). Furthermore, feature importance also can explain why/how the prediction is given (questions 47, 50). To answer  question 49 \textit{Why are instance A and instance B given the same prediction?}, we show one  explanation per instance. SHAP can explain both, image and tabular data and the feature importance values have been shown to be consistent with human judgment. We, therefore, use SHAP to answer the questions that require feature importance information. 

DICE~\cite{mothilal_explaining_2020} is an explanation method, focusing on counterfactual explanations for tabular data. Given an instance, DICE searches for the minimum feature changes required to get a different prediction, and therefore provides information to answer questions 12, 14, and 51--53. DICE can also identify required changes for a specific feature to change the prediction to a different target class and therefore provides the information for answering question 13: \textit{How should this feature change to get a different prediction?}. \footnote{Despite limitations of DICE in generating actionable counterfactual explanations~\cite{guidotti2022counterfactual}, we include this method in our study due to its alignment with our predefined criteria and high overall quality\cite{moreira_benchmarking_2022, guidotti2022counterfactual}.}
Similar to DICE, CFProto~\cite{van_looveren_interpretable_2020} is a counterfactual explanation method applicable to image data. However, the method does not allow to change single, specific features to obtain a target class, because features in image data are hard to define. Thus, we apply CFProto to gather information for answering questions 12, 14, and 51--53 on image data.
Anchor~\cite{ribeiro_anchors_2018} computes sufficient conditions for a prediction, so called anchors, such that as long as the anchor holds, changes to the remaining feature values of the instance do not matter. Therefore, it can be used to determine the boundaries of the prediction, which are suitable to answer questions 17--20.


In summary, to integrate XAI into the dialogue policy component, we systematically map  XAI questions to XAI methods. 

\section{Generating Answers}
\label{sec:approach:nlg}
XAI methods provide the core information to answer the corresponding questions, but they lack  explanatory text in natural language for the end user.  
Presenting just the raw information in form of a table or importance values alongside feature names to end users is not always adequate. Instead, additional context, such as what the values represent or how to interpret them, is desirable.
To address this problem, we incorporate a template-based natural language generation component. For each question, we define text templates (partially with dataset-specific vocabulary) containing placeholders for the information obtained from XAI methods. We combine this information with -- or convert it to -- textual explanations depending on the type of information generated by the XAI method. For images and graphs (e.g. SHAP's outputs), we add a textual explanation. For tabular format (e.g. DICE's outputs), we convert the table to natural language by extracting feature names and corresponding values.
For example, in Figure~\ref{fig:result:example_adult}, the answer ``The Occupation should be changed to Tech-support to get an income of more than 50K'' is the combination of the template ``The $<$feature$>$ should be changed to $<$value$>$ to get $<$class$>$'' and the counterfactual information obtained from DICE~\cite{mothilal_explaining_2020}.

In case of counterfactual explanations, relations need to be extracted in addition to feature names and values. 
For the counterfactual question 53 (\textit{Why is this instance predicted P instead of Q?}), we extract and compare the relation between the given instance with class P and a counterfactual instance with the target class Q. In particular, for numerical features, we identify the relation between two values, whether the first is greater or smaller than the second. In the case of categorical/textual features, we check if a value is different from the other. We then use the corresponding templates for the extracted relations.  For example, in Figure~\ref{fig:result:example_adult}, we identify 2 relations between the given instance (label \textit{\textless=50K}) and its counterfactual (label \textit{\textgreater50K}): the numerical feature ``Age = 39'' is smaller than ``Age = 66.3'' and the categorical feature ``Workclass = State-gov'' is different to ``Workclass = Self-emp-inc''. The corresponding templates  are ``$<$feature$>$ is too low'', and ``$<$feature$>$ is not suitable''. 
The next sentence ``If Age increases to at least 66.3 and Workclass changes to ...'' also contains 2 different templates for these relations. 
Finally, we combine the templates with the XAI information to return the answers. 

For questions that cannot be answered (either below the matching threshold or no available methods), we define a template asking users for an alternative phrasing or a different question.   


\section{Evaluation of NLU Component}
\label{sec:eval}
In this section, we describe the evaluation of promising approaches to implement the natural language understanding (NLU) component, i.e., the mapping from user input to XAI questions as described in section~\ref{ssec:approach:nlu}. Specifically, we compare traditional text classification to sentence embeddings.

\subsection{Experimental Setup}
We use the XAI question phrase bank as described in Section~\ref{sec:question-phrase-bank}, i.e., the data set with reference questions and paraphrases after manual quality control.
We use the reference question ID as a label and assign a common label to sets of questions with identical answers in the same category. 
For instance, question~2
\textit{How does feature $f$ impact its predictions?} and question~5 \textit{Is feature $f$ used or not used for the predictions?} both ask for (binary) feature contribution of one specific feature and can be answered with the feature importance information for feature f of an XAI method. 
After relabeling, the final data set contains 329 instances and 52 different labels (from 73 initial questions).
Additionally, we also evaluate our models on the subset of XAI questions only, i.e., 23 questions whose rows are highlighted in gray in~\cref{tab:approach:xai-question-and-methods}. This XAI-only set contains 111 instances and 14 labels. 

We compare four different approaches using two different feature representation methods: classical TF-IDF weighting, and sentence embeddings extracted by SimCSE~\cite{gao_simcse_2021} (More details of representation methods can be found in the Appendix, Section~\ref{sec:append:representation}) . On the TF-IDF vector space, we evaluate two classifiers commonly applied to text classification:  Support Vector Machines (SVMs) with various kernels and Random Forests (RF).
On the sentence embedding space (SimCSE), we compare a similarity-based approach to a supervised model. 
In the similarity-based approach, we rank reference questions by their cosine similarity to a user question. As a supervised model, we use a fully connected feedforward neural network (NN) with a single hidden layer of size 256, trained with cross-entropy loss.
We employ grid-search for hyperparameter tuning with details available in the Appendix, Table~\ref{tab:append:nlu-hyper}. 

We present the mean of micro- and macro-averaged F1 scores on 3-fold cross-validation. For multi-class classification, the micro-averaged F1 score is equal to accuracy. To further evaluate the importance of paraphrases, we compare all approaches without training on additional paraphrases. Specifically, for SVM, RF and SimCSE + NN, we train the models on only 73 reference questions (\cref{tab:approach:xai-question-and-methods}) and evaluate them on the generated paraphrases.

\subsection{Results}
\begin{table}[tbp]
\centering
\setlength\tabcolsep{1pt}
\caption{Evaluation results for 3-fold cross-validation. Showing mean on the full data set ($std <=0.05$) and on the subset of XAI questions ($std <= 0.09$) with (and without) training on paraphrases. 
}
\label{tab:result:nlu-methods}
\scriptsize
\begin{tabular}{+l^l^l^l^l}
\toprule
 & \multicolumn{2}{c}{\textbf{All questions}} & \multicolumn{2}{c}{\textbf{XAI questions only}}\\
 & \textbf{Accuracy} & \textbf{Macro F1} & \textbf{Accuracy} & \textbf{Macro F1}\\\otoprule
SVM + TF-IDF & $0.61\pm 0.05$($0.55$)  & $0.50\pm 0.04$($0.49$) & $0.67\pm 0.05$($0.62$)&  $0.59\pm 0.06$($0.53$)\\
RF + TF-IDF & $0.57\pm 0.04$($0.39$)  & $0.47\pm 0.05$($0.30$) & $0.68\pm 0.05$($0.53$)&  $0.55\pm 0.06$($0.45$)\\
SimCSE + Cosine & \textbf{0.72} $\pm$ \textbf{0.03}($0.61$)& $0.65\pm0.05$($0.57$)& $0.83\pm0.03$($0.58$)&  $0.78\pm0.06$($0.52$)\\
SimCSE + NN & \textbf{0.72} $\pm$ \textbf{0.03}($0.55$) & \textbf{0.67} $\pm$ \textbf{0.03} ($0.45$)& \textbf{0.85} $\pm$ \textbf{0.08}($0.67$)&  \textbf{0.83} $\pm$ \textbf{0.09}($0.61$)\\\bottomrule
\end{tabular}
%
\end{table}
\cref{tab:result:nlu-methods} shows the evaluation results (more details can be found in Appendix, Section~\ref{apdx:nlu-details}). 
The supervised approach based on sentence embeddings (SimCSE + NN) outperforms the other approaches on both, the full data set and the XAI subset with an accuracy of 0.85 on the XAI subset. Both traditional text classification approaches (RF + TF-IDF and SVM + TF-IDF) are already outperformed by the unsupervised similarity-based approach on sentence embeddings (SimCSE + Cosine) and even more so by the supervised approach on sentence embeddings (SimCSE + NN), highlighting the efficiency of pre-trained models for natural language processing tasks. In addition, for all approaches, training with paraphrases yields significantly higher results than without paraphrases indicating the strong impact of the generated paraphrases.


\section{Conclusion}
\label{sec:conclusion}
Following the conversational style of human-to-human explanations, we leveraged a conversational agent to explain machine learning models. To capture the variance of questions that can be asked about the topic, we extended an XAI question bank with paraphrases. Each question-paraphrases set defines a specific information need, represented by a reference question. We presented a systematic analysis of methods that can address those information needs aiming at a sufficient, but small subset of all available XAI methods. Our XAI question phrase bank and XAI method collection can serve as guidance for the future development of conversational XAI agents. In future work, we plan to integrate a learning component for dialogue policies to make the system self-adaptable from interactions. Furthermore, it is essential to extensively evaluate the framework through human evaluation and diverse datasets, which we leave for future work.

\clearpage
\bibliographystyle{splncs04}
\bibliography{References}
\clearpage\section*{Appendix}
\appendix
\section{GPT-3 Paraphrase  Prompting}
\label{asec:gpt-3-paraphrases}
\begin{figure}[H]
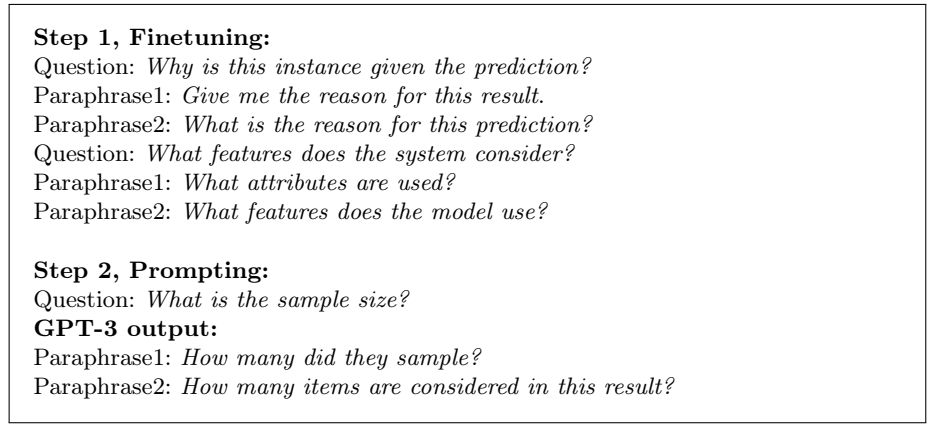

    \begin{framed}
    \small
    \noindent \textbf{Step 1, Finetuning:}\\
   Question:  \textit{Why is this instance given the prediction?} \\
    Paraphrase1: \textit{Give me the reason for this result}.\\
    Paraphrase2: \textit{What is the reason for this prediction?}\\
    Question: \textit{What features does the system consider?}\\
    Paraphrase1: \textit{What attributes are used?}\\
    Paraphrase2: \textit{What features does the model use?}\\
    
    \noindent\textbf{Step 2, Prompting:}\\
    Question: \textit{What is the sample size?\\}
    \noindent\textbf{GPT-3 output:}\\
    Paraphrase1: \textit{How many did they sample?}\\
    Paraphrase2: \textit{How many items are considered in this 
    result? }
    \end{framed}
\caption{Example GPT-3 finetuning, prompt and output to generate XAI paraphrase candidates}
\label{fig:prompt}
\end{figure}

We finetune the GPT-3 model with two instances for each reference question in the initial XAI question bank (2-shot).
Each instance consists of the reference question and two paraphrases of this question.
Subsequently, we prompt the model to generate paraphrases with a new question (see Fig.~\ref{fig:prompt} for an example). We repeat the prompt multiple times for each reference question.

\section{Phrase Annotation Details}
\label{sec:append:annotation}
\begin{figure}[h]
    \centering
    \includegraphics[width=0.8\textwidth]{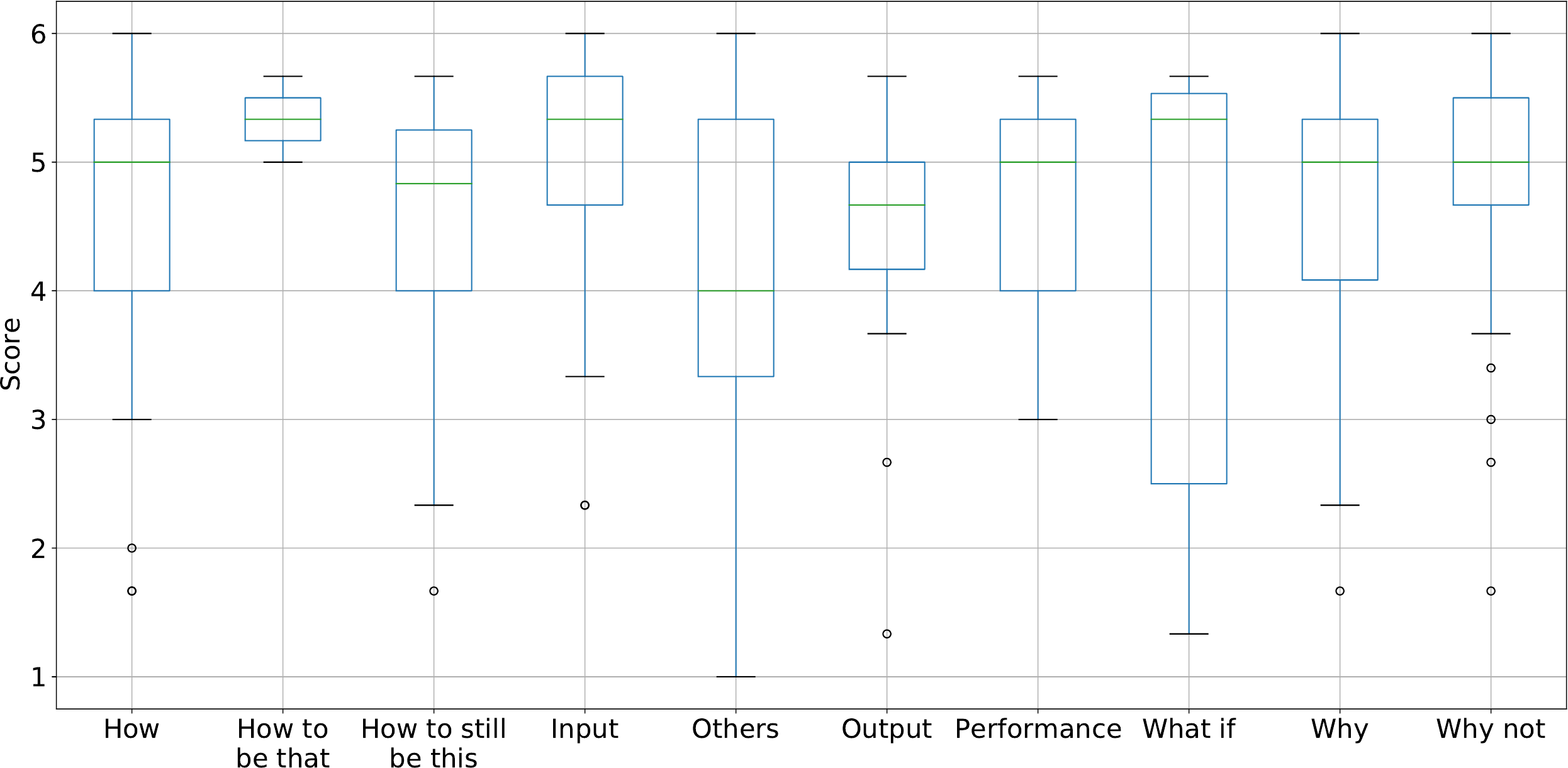}
    \caption{Annotation score distribution for each question category.}
    \label{fig:dataset:box-plot-scores}
\end{figure}

The distribution of annotation scores varies among each question category (see Fig.~\ref{fig:dataset:box-plot-scores}). In general, most of the score medians are above 4, indicating the good quality of GPT-3 in generating paraphrases. However, varying interquartile ranges suggest that GPT-3 generates better paraphrases in specific categories such as \textit{How to be that} or \textit{Why not}, and mixed paraphrases in others, such as \textit{What if} or \textit{Other}.

\begin{figure}[h]
    \centering
    \includegraphics[width=0.8\linewidth]{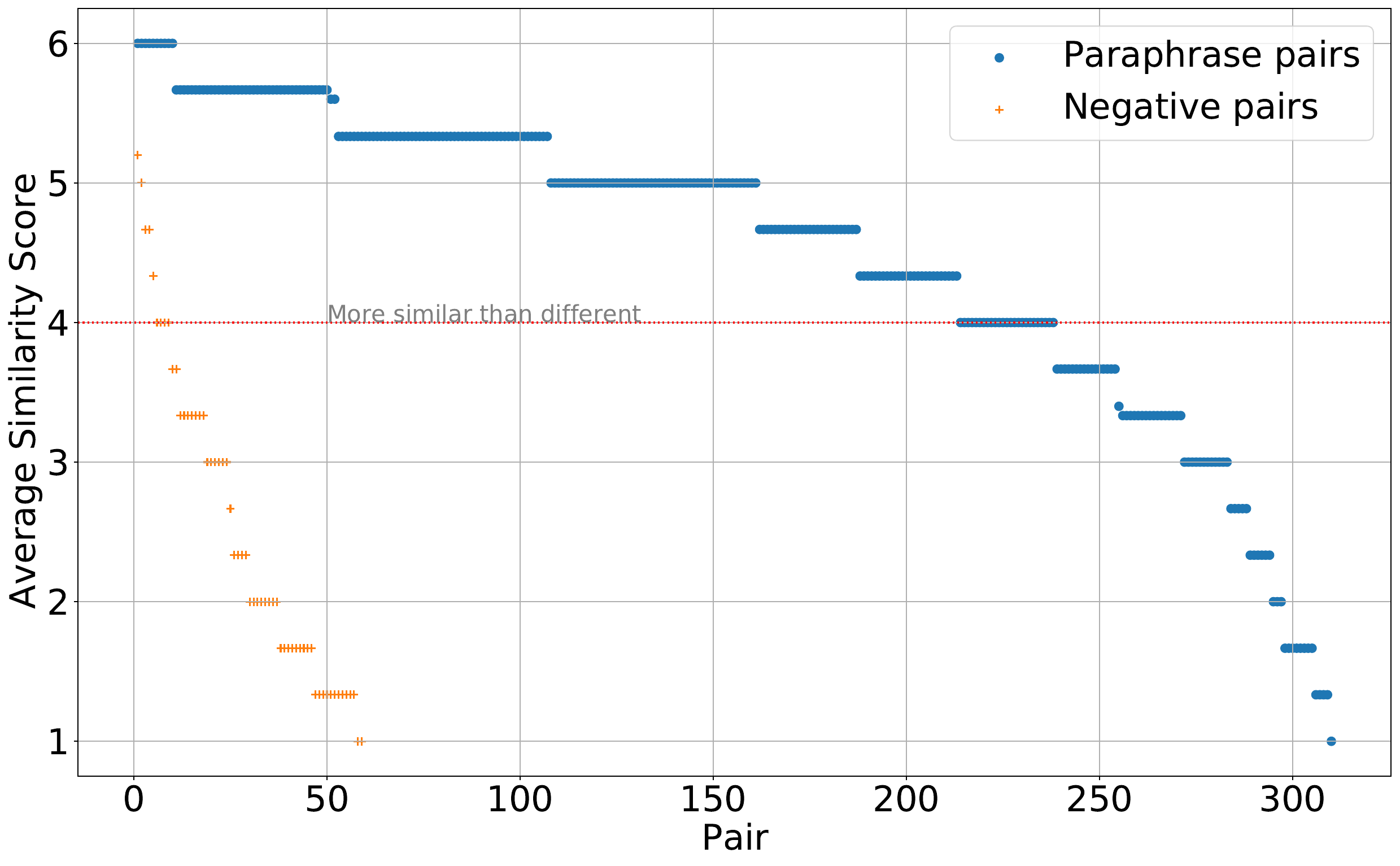}
    \caption{Average human annotation score for all phrase pairs ranked by score. Negative pairs are paraphrases from different questions.}
    \label{fig:dataset:phrase-pair-scores}
\end{figure}

Fig.~\ref{fig:dataset:phrase-pair-scores} depicts the average annotator score per phrase pair. 
Phrase pairs are ranked by their score, separately for the 310 paraphrase pairs and 59 negative pairs. Most of the paraphrase pairs that were generated by GPT-3 have a score $\geq 4$, and thus are perceived as being similar, indicating that GPT-3 generates high-quality paraphrases in general. Conversely, most negative pairs, which were sampled from different questions, have an average score  $<4$, supporting the quality of the human annotations. 
However, there are a few outliers of negative pairs which are annotated with a high similarity score. This is likely caused by our choice of negative phrases, sampled at random from a different question. These pairs may not be truly negative, as one question may be more general than the other or they can be interpreted in different ways (see Table~\ref{tab:append:negative-pairs-high-score} for examples).
Furthermore, annotators disagree on ambiguous pairs and agree on unambiguous pairs (Table~\ref{tab:append:annotation-pairs}), further supporting the good quality of the dataset.

\section{Representation Methods}
\label{sec:append:representation}
We test two different feature representation methods: classical TF-IDF weighting, and sentence embeddings. 
For TF-IDF weighting, we follow a standard preprocessing pipeline: We select tokens of 2 or more alphanumeric characters (punctuation is ignored and always treated as a token separator) and stem the text using the Porter Stemmer~\cite{Manning2008_introduction-to-IR} to obtain our token dictionary. Maximum and minimum DF thresholds are subject to hyperparameter optimization (see full list of hyperparameters in Table~\ref{tab:append:nlu-hyper}).
We embed sentences (i.e., question instances) using SimCSE~\cite{gao_simcse_2021} to obtain an alternative feature representation to TF-IDF. We employ the pretrained RoBERTa-large model~\cite{liu_roberta_2019} as base model in SimCSE.

\begin{table*}[t]
\footnotesize
\caption{Example negative pairs with average score $>$ 4}
\label{tab:append:negative-pairs-high-score}
\begin{tabular}{+p{5.5cm}^p{5.5cm}^c}
\toprule \tabhead
Phrase A & Phrase B & Scores \\ \otoprule
How are the predictions made?
 & What kind of algorithm is used? & 5,5,4\\[2pt]
How should this Instance change to get a different prediction?
 & What should be the value of this feature in order to change the prediction?& 5,5,4\\[2pt]
What features of this instance lead to the system's prediction?	& How was this instance given this value/category?& 5,5,3\\
\bottomrule
\end{tabular}
\end{table*}
\begin{table*}[h]
\footnotesize
\caption{Phrase pairs with highest agreement/disagreement between annotators (bold indicates the reference questions in the question bank)}
\label{tab:append:annotation-pairs}
\begin{tabular}{+p{5.5cm}^p{5.5cm}^c}
\toprule \tabhead
Phrase A & Phrase B & Scores \\ \otoprule
\multicolumn{3}{c}{\textbf{Disagreement}} \\[3pt]
Why this instance has class P but Q does not?
 & \textbf{Why is this instance predicted P instead of Q?}&6,4,1\\[2pt]
\textbf{How much data like this is the system trained on?}
 & Give me an instance which is similar to this.
&5,1,1\\[2pt]
\textbf{How is this instance not predicted A?}& How is the result B for this instance possible? &6,1,2\\\midrule
\multicolumn{3}{c}{\textbf{Agreement}} \\[3pt]
Which features does it take into account?
 & \textbf{What features does the system consider?}&6,6,6\\[2pt]
\textbf{What are the results of other people using the system?}
 & What was the result when other people used the system?
&5,5,5\\[2pt]
What is the reason for this prediction?
& Which features does the system use? &2,2,2\\
\bottomrule
\end{tabular}
\end{table*}


\begin{table}[t]
\scriptsize
\centering
\caption{Hyperparameters for Grid Search, \textbf{bold} indicates the chosen hyperparameters. For the other hyperparameters, we use default value in scikit-learn~\cite{scikit-learn}.}
\label{tab:append:nlu-hyper}
\begin{tabular}{+ll}
\toprule \tabhead
Model & \textbf{Hyperparameter} \\ \otoprule
TF-IDF & max\_df=[1.0,\textbf{0.8}]; min\_df=[0.1,0.2,\textbf{1}]\\[2pt]
SVM & kernel=['linear', 'poly', \textbf{'rbf'}, 'sigmoid', 'precomputed'];C=[0.1, 1, 10, \textbf{100}, 1000]; \\ &gamma=[\textbf{0.1}, 1, 10, 100];degree=[0,1,2,\textbf{3}]\\
RF & bootstrap=[\textbf{True}, False]; max\_depth=[10, \textbf{None}];min\_samples\_leaf=[1, \textbf{2}, 4]; \\
    &min\_samples\_split=[\textbf{2}, 5, 10];n\_estimators=[10, \textbf{50}, 100]\\
NN & Epoch=[\textbf{50}, 100]; batch\_size=\textbf{1}; learning\_rate=[\textbf{4}, 6]\\
\bottomrule
\end{tabular}
\end{table}

 \begin{figure}[H]
    \includegraphics[width=\linewidth]{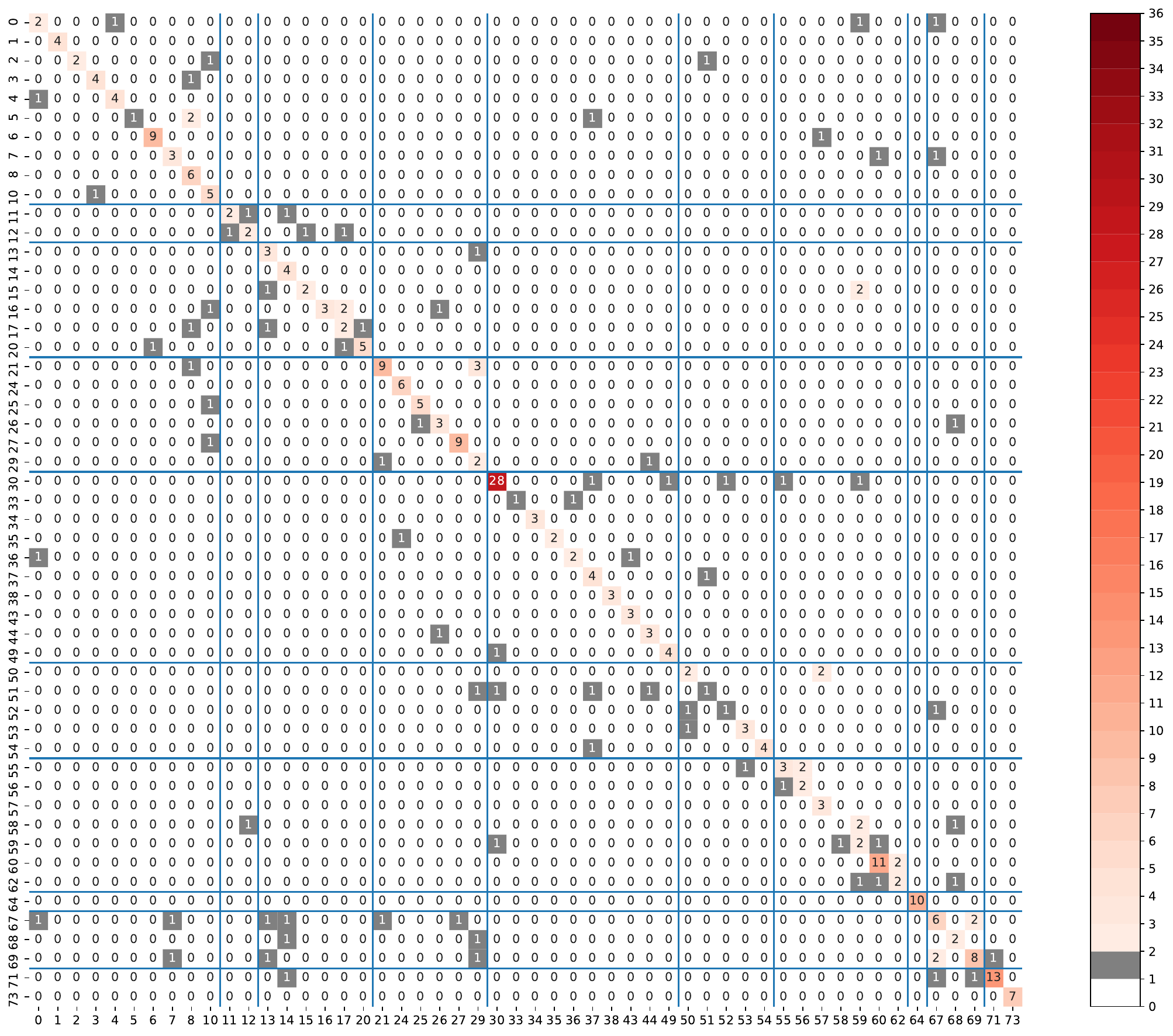}
    \caption{Confusion matrix for SimCSE + NN}
    \label{fig:appdx:nlu-eval:confusion-matrix}
\end{figure}

\section{Details on NLU Evaluation }
\label{apdx:nlu-details}



Figure~\ref{fig:appdx:nlu-eval:confusion-matrix} shows the confusion matrix for SimCSE + NN's. The blue lines separate the questions in each category (see Table~\ref{tab:approach:xai-question-and-methods}), and the diagonal contains number of the True Positive rate for each question. This prominent diagonal reflects the high accuracy of the approaches. The squares around the diagonal are sub-confusion matrices between questions in the same group. The high number of gray color in these squares indicates that questions in the same category are harder to distinguish than questions in different category (note that the numbers on x and y axes indicate the merged labels, not IDs).

\section{XAI Method Overview}
 \label{apdx:analyis-xai-methods-libraries}

 \begin{table*}[t]
\centering
\scriptsize
\caption{XAI methods and selection criteria (Abbreviation: Cls = Classification, Reg = Regression, RL = Reinforcement Learning)}
\label{tab:apdx:xai-method-and-questions}
\begin{tabular}{+p{2.5cm}^l^P{1.5cm}^c^p{2cm}^p{2cm}^l}
\toprule \tabhead
Method& Year& Reference Question  & Code  & Type of data & Model & Problem  \\\otoprule
GAM~\cite{hastie2017generalized}&1990 &2-9,11,47-50,68-73& &Tabular&Intrinsically interpretable&Cls, Reg\\
GoldenEye~\cite{henelius2014peek}&2014 &5,8&\checkmark(in R)&Tabular&All&Cls\\ 
MFV~\cite{nguyen2016multifaceted}&2016 &4&\checkmark&Image&Neural Network&Cls\\ 
LIME~\cite{ribeiro_why_2016} &2016 &2,3,5,6,8,47-50&\checkmark&Tabular, Image, Text&All&Cls, Reg\\ 
Feature Tweaking~\cite{tolomei2017interpretable} &2017&2,3,6,12&\checkmark&Tabular&Tree-based&Cls\\ 
Model Extraction~\cite{bastani2017interpretability} &2017&2-9,11,47-50,68-73& &Tabular&All&Cls, Reg, RL\\ 
SHAP~\cite{lundberg_unified_2017}&2017 &2,3,5,6,8,47-50&\checkmark&Tabular, Image, Text&All&Cls, Reg\\ 
ANCHOR~\cite{ribeiro_anchors_2018}&2018 &17-20&\checkmark& Tabular, Text&All &Cls\\
Boolean rules~\cite{dash2018boolean}&2018 &2-9,11,47-50,68-73&\checkmark&Tabular&Intrinsically interpretable&Cls\\ 
LORE~\cite{guidotti2018local}&2018 &2,3,5,6,8,47-50&\checkmark&Tabular&All&Cls\\
CEM~\cite{dhurandhar_explanations_2018}&2018 &12,14,51-53&\checkmark& Image &Neural Network &Cls\\
ProtoPNet~\cite{chen2019looks}&2019 &2-9,11,47-50,68-73&\checkmark&Image&Neural Network&Cls\\
DICE~\cite{mothilal_explaining_2020}&2020 &12-14,51-53&\checkmark& Tabular &All &Cls\\
CFProto~\cite{van_looveren_interpretable_2020}&2020 &12,14,51-53&\checkmark& Tabular, Image &All &Cls\\
ProtoTree~\cite{nauta2021neural}&2021 &2-9,11,47-50,68-73&\checkmark& Image&Neural Network&Cls\\
\bottomrule
\end{tabular}
\end{table*}
Table~\ref{tab:apdx:xai-method-and-questions} shows the criteria, which are mentioned in section 5.2 in the paper, to choose the proper XAI method for each XAI question.

\section{Conversation Scenarios}
\label{sec:scenarios}
\subsection{Random Forest Classifier on Adult Data}
In this section, we show an example conversation between a prototype implementation of our proposed framework and a user on tabular data (Adult dataset\footnote{\url{https://archive.ics.uci.edu/ml/datasets/adult/}})  with a Random Forest (RF) classifier. 

The task on this data set is to predict whether the income exceeds \$50.000/year (abbreviated 50K) based on census data. 
We train the classifier using the sklearn library  and its standard  parameter settings\footnote{\url{https://scikit-learn.org/stable/modules/generated/sklearn.ensemble.RandomForestClassifier.html}}. The mean accuracy of the classifier using 3-fold cross-validation is 0.85. For explanations, we retrain the RF classifier with the same parameter settings on the full data set. The data set and the classifier are loaded at the beginning of the conversation. 

Figure 1 in the main body of the paper shows a conversation with the prototype agent (X-Agent). At the beginning of the conversation, the user provides information about her features by answering retrieval questions from the agent. These questions can be generated based on DataSheets~\cite{Gebru2021_commacm_datasheets} of the data set. We omit this part of the conversation in 
Fig.~\ref{fig:result:example_adult}
and show how the X-Agent reacts to several questions about the model. 


The first question is the request: \textit{Give me the reason for this prediction!} The natural language understanding (NLU) component matches this question to the reference question \textit{Why is this instance given this prediction?} in the question bank (question 47 in Table~\ref{tab:approach:xai-question-and-methods}).
The Question-XAI method mapping (QX) selects SHAP~\cite{lundberg_unified_2017} as the XAI method to provide the information for the answer. The natural language generation (NLG) component combines SHAP's feature importance information with the predefined text ``The above graph ...'' to respond to the user question. 

For the next question, \textit{Why is this profile predicted \textless=50K instead of \textgreater50K}, the labels \textit{\textless=50K} and \textit{\textgreater50K} are replaced by the token $<$class$>$ before matching to reference question 53 in
Table 2 (main body of the paper)
\textit{Why is this instance predicted P instead of Q?}. The QX component identifies  DICE~\cite{mothilal_explaining_2020} as the explanation method for this reference question, and the information is translated into natural language. 
In detail, DICE returns a counterfactual instance with the desired target label (\textit{\textgreater50K}), yielding two features (Age and Workclass) that need to change in order to obtain the desired prediction.
The NLG component extracts the relations between feature values of the original instance (Age:39, Workclass:State-gov) and counterfactual instance (Age:66.3, Workclass:Self-emp-inc). In comparison to the counterfactual, Age of the original instance is lower and Workclass differs. 
These relations are converted and rendered as text in the final answer by the NLG component.

For the final question \textit{That's hard, how could I change only Occupation to get \textgreater50K prediction?}, the words ``Occupation'' and ``\textgreater50K'' are substituted by tokens $<$feature$>$ and $<$class$>$ respectively. Then, the question is matched to reference question 13 
(see Table~\ref{tab:approach:xai-question-and-methods})
\textit{How should this feature change to get a different prediction?}.  DICE is again determined as the  XAI method for providing the required information to answer this question. 
However, this question asks for a specific feature, i.e., constrains the search space of DICE for counterfactuals. Finally, the provided  information is again translated to natural language.
\begin{figure}[ht!]
\footnotesize{\ttfamily
\textcolor{ansi-blue-intense}{\textbf{X-Agent: }}This is your input image.\\
    \includegraphics[width=0.2\linewidth]{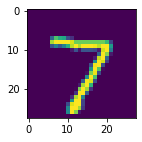}\\
My prediction for this image is number 7.\\[2pt]
\textcolor{ansi-red-intense}{\textbf{User:}} Why did you predict that?\\
\textcolor{ansi-blue-intense}{\textbf{X-Agent: }}\\
    \includegraphics[width=0.2\linewidth]{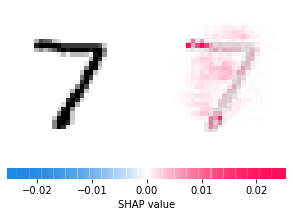}

On the right image above, I highlight the important parts (red color) of the image that impact this prediction.\\[2pt]
\textcolor{ansi-red-intense}{\textbf{User:}} How should this image change to get number 9 prediction?\\
\textcolor{ansi-blue-intense}{\textbf{X-Agent: }}\\
   \includegraphics[width=0.2\linewidth]{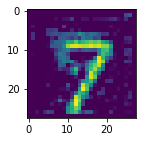}\\
Here you go! I just modified your image a bit to make it look like number 9.
        \caption{Conversation example to explain a  Convolutional Neural Network on MNIST}
    \label{fig:result:example_mnist}
}
\end{figure}

\subsection{Convolutional Neural Network on MNIST}
\label{sec:scenarios:cnn-on-mnist}

We use the MNIST data set and a pre-trained convolutional neural network~\cite{van_looveren_interpretable_2020} to showcase a conversation on an image data set (see Fig.~\ref{fig:result:example_mnist}). 
First, the NLU component matches the first question \textit{Why did you predict that?} to reference question 47 \textit{Why is this instance given this prediction?} (see Table~\ref{tab:approach:xai-question-and-methods} ). Then, QX maps this question to SHAP~\cite{lundberg_unified_2017} as the explanation technique. SHAP highlights the important parts on the image that lead to prediction 7. The NLG component adds an explanation in form of natural language text to the information provided by SHAP (the image). For the second question \textit{How should this image change to get number 9 predicted?}, \textit{number 9} is replaced by token $<$class$>$. NLU maps this processed question to reference question 12 (see Table~\ref{tab:approach:xai-question-and-methods}). QX identifies CFProto~\cite{van_looveren_interpretable_2020} as the method to answer this question. CFProto outputs the modified image that is closer to number 9. Finally, NLG generates the explanation text along with the output of CFProto.
%
%
%

%




\end{document}